\documentclass{article} 
\usepackage{preprint,times}

\usepackage{amsmath,amsfonts,bm}

\def\eqref#1{equation~\ref{#1}}

\def\1{\bm{1}}

\DeclareMathAlphabet{\mathsfit}{\encodingdefault}{\sfdefault}{m}{sl}
\SetMathAlphabet{\mathsfit}{bold}{\encodingdefault}{\sfdefault}{bx}{n}

\newcommand{\Var}{\mathrm{Var}}

\newcommand{\Cov}{\mathrm{Cov}}

\usepackage{hyperref}
\usepackage{url}
\usepackage{amsmath,amssymb,amsthm}
\usepackage{booktabs}
\usepackage{multirow}
\usepackage{graphicx}
\usepackage{xcolor}
\usepackage{listings}
\usepackage{mathtools}

\newtheorem{proposition}{Proposition}
\newtheorem{corollary}{Corollary}
\newtheorem{remark}{Remark}

\newcommand{\sdot}{\dot{s}}
\newcommand{\mdot}{\dot{m}}
\newcommand{\vdot}{\dot{v}}
\newcommand{\Ex}{\mathbb{E}}
\newcommand{\gpcfm}{GP-CFM}
\newcommand{\icfm}{I-CFM}
\newcommand{\Wtwo}{W_2}

\title{What Does a Stream Model Buy You in Flow Matching?}

\author{Jian Xu\\
RIKEN\\
\texttt{jian.xu@riken.jp}}

\finalcopy
\begin{document}

\maketitle

\begin{abstract}
Stream-level flow matching (\gpcfm; \citealp{wei2025stream}) replaces the linear interpolant of conditional flow matching (CFM) by a Gaussian-process (GP) \emph{stream} connecting each source--target pair, and reports lower sample error than \icfm{} on 2-Gaussian, MNIST and CIFAR-10 benchmarks.
We ask what such a stream model actually contributes.
Three results answer the question.
(i)~\emph{Reduction.} The stream-level CFM objective depends on the stream law only through the per-time joint law of $(s_t,\sdot_t)$, so the conditional paths a Gaussian stream can reach are exactly the Gaussian conditional paths CFM already parametrises; in the coordinate-wise, shared-scalar-kernel construction \gpcfm{} actually uses, the entire design space collapses to two scalar curves $(m_t,v_t)$, and cross-time covariance affects only estimator variance.
(ii)~\emph{The GP is a constrained chart of that space.} One kernel sets both $m_t$ and $v_t$, so the paper's own recipe for widening coverage---shrinking the SE length-scale---destroys the interpolant (the midpoint mean weight falls from $1.03$ to $0.00$). On the 2-Gaussian benchmark this makes the GP chart diverge on $15/200$ runs at high coverage against $0/200$ for a decoupled $(m_t,v_t)$ chart ($p=6.6\times10^{-5}$), and crossing the two curves shows the divergence tracks the mean, not the variance. On MNIST the same sweep does not diverge and the ordering reverses, so whether the coupling is harmful is benchmark-dependent; what holds on both is that the recipe buys nothing---no coverage level beats the paper's own, and past $\max_t\sqrt{v_t}\approx0.6$ both charts degrade.
(iii)~\emph{Audit.} The released code does not implement the mechanism it describes: state and velocity are drawn independently ($\mathrm{corr}=0.00\pm0.01$ against an intended $\pm0.83$--$0.99$), with variances taken from misindexed covariance entries. Independently of the bug, the flagship 2-Gaussian table has no power (the reported $0.03$ $\Wtwo$ gap needs ${\approx}7{,}600$ seeds; $100$ were run) and in the reported configuration the GP contributes $0.4\%$ state spread and $0.12\%$ velocity noise. On MNIST the released method is \emph{worse} than \icfm{} by $2.6$ FID (paired, $p<10^{-3}$) and on CIFAR-10 the released sampler is worse by $1.7$ FID on every seed; correct implementations of the intended law are equivalent to \icfm{} within the paper's own claimed effect, and fixing the bug recovers $2.2$--$2.3$ FID on MNIST.
For the streams \gpcfm{} builds, the design space is two curves, and a GP is a constrained way to write them down---not a richer latent.
\end{abstract}

\section{Introduction}
\label{sec:intro}

Conditional flow matching (CFM) trains a velocity field $v_\theta(t,x)$ by regressing on a \emph{conditional} velocity $u_t(x\mid z)$ whose expectation over the latent $z$ is the marginal field that transports source to target \citep{lipman2023flow,tong2024improving,albergo2023building}. The choice of $z$ and of the conditional path $p_t(\cdot\mid z)$ is the main design freedom of the framework.
Stream-level flow matching \citep{wei2025stream} (henceforth \gpcfm) proposes to go ``one level deeper'': condition on an entire latent path (a \emph{stream}) $s=\{s_t\}_{t\in[0,1]}$ joining $x_0$ to $x_1$, endow it with a Gaussian-process (GP) law, and regress on the stream velocity $\sdot_t$. The motivation is that GP streams spread the training distribution over $(t,x)$ space, smoothing the target field and reducing the variance of its estimate; the paper reports improvements over the linear interpolant (\icfm) on a 2-Gaussian toy, MNIST and CIFAR-10, and the approach was accepted at ICML 2025.

This paper asks a narrow question: \emph{what does the stream model contribute that the conditional path does not already provide?} The answer is ``nothing beyond two scalar curves, and it costs you a failure mode''. We reach it in three steps.

\paragraph{Reduction (Section~\ref{sec:reduction}).}
The stream-level objective is $\Ex_{t,s}\|v_\theta(t,s_t)-\sdot_t\|^2$. It depends on the stream law only through the family of per-time joint laws $\{P(s_t,\sdot_t\mid z)\}_t$. For a Gaussian stream, mean-square differentiability ties the state--velocity covariance to the variance curve, $\dot V_t=C_t+C_t^\top$, so the conditional path is $\mathcal N(m_t,V_t)$ and the regression target is its CFM field \citep{lipman2023flow} up to a divergence-free rotational term. The reachable set of conditional paths is therefore exactly the Gaussian conditional paths CFM already parametrises; in the coordinate-wise, shared-scalar-kernel construction \gpcfm{} uses, the rotational term vanishes and the design space is the two scalar curves $(m_t,v_t)$. What is left over is inert: residual velocity noise cannot move the minimiser and only inflates gradient variance, and cross-time covariance---the very thing that makes a stream a stream---enters only through the correlation of multiple time draws within a minibatch.

\paragraph{The GP is a constrained chart (Section~\ref{sec:chart}).}
The reduction says the design space is $(m_t,v_t)$. In \gpcfm{} both curves come from one squared-exponential (SE) kernel conditioned on the endpoints: the mean is the SE interpolant and the variance the SE bridge variance. The paper's recipe for wider coverage is to shrink the length-scale. We show analytically that this cannot be done without destroying the interpolant: as $\ell\to0$ the midpoint weight $a_0(\tfrac12)+a_1(\tfrac12)$ on the endpoints falls from $1.03$ ($\ell=2$) to $0.27$ ($\ell=0.25$) to $0.00$ ($\ell=0.1$), i.e.\ the ``stream'' between $x_0$ and $x_1$ detours through the prior mean. Coverage sweeps on the paper's own benchmarks make the consequence measurable, and it is not uniform. On the 2-Gaussian toy, beyond a state spread of $\approx0.6$ the GP chart diverges on $12$--$18\%$ of training runs against $0/200$ for a decoupled chart, and crossing the two curves pins the divergence on the mean. On MNIST neither chart diverges and the GP chart is the better of the two, so we report the cliff as a toy-specific consequence of a coupling that is itself analytic. On both benchmarks, however, the recipe fails to pay: no coverage level is significantly better than the one the paper ships.

\paragraph{Audit (Section~\ref{sec:audit}).}
The released sampler does not implement the joint law it derives: a \texttt{view} where a transpose was needed pairs each stream's mean with another stream's conditional covariance and reads state and velocity from different rows, so the two are drawn \emph{independently}. We verify the mechanism cell by cell; the bug is present in five of the seven released notebooks, including the MNIST one. Independently of the bug, the flagship 2-Gaussian gap ($1.54$ vs $1.51$ $\Wtwo$, SE $0.08$) is $0.3$ standard errors: detecting it at $80\%$ power would take ${\approx}7{,}600$ seeds, not $100$. On MNIST the released method is worse than \icfm{} by $2.56$ FID ($1/20$ paired wins), every correct implementation of the intended law is equivalent to \icfm{} within the effect the paper claims (TOST at a $0.95$ FID margin, $p\le0.006$), and fixing the bug recovers $2.2$--$2.3$ FID.

\paragraph{Contributions.}
(1) A reduction theorem: stream-level CFM with Gaussian streams reaches exactly the Gaussian conditional paths, with the stream's cross-time structure inert in the objective (Propositions~\ref{prop:reduction}--\ref{prop:gaussian}); for the construction \gpcfm{} uses this is two scalar curves (Corollary~\ref{cor:scalar}).
(2) A structural constraint of the GP parametrisation: mean and variance are coupled through one kernel, so the paper's own prescription for coverage collapses the mean (Proposition~\ref{prop:chart}). We measure its consequences on both benchmarks---a $15/200$ vs $0/200$ divergence gap on the toy, attributed to the mean curve by a $2\times2$ crossing, and no divergence but no benefit on MNIST.
(3) A reproduction audit of \gpcfm: the released sampler draws state and velocity independently; the toy table has no power; the MNIST gain reverses sign under a paired protocol and the CIFAR-10 one is not reproduced; the bug costs $1.7$--$2.3$ FID.
We are careful about what the positive result does \emph{not} say: no arm beats any other in median quality on any benchmark we ran, and larger coverage does not improve the 2-Gaussian benchmark. The claim is controllability and cost, not accuracy.

\section{Background}
\label{sec:background}

\paragraph{Conditional flow matching.}
Let $q_0,q_1$ be source and target on $\mathbb R^d$ and $\pi$ a coupling. CFM \citep{lipman2023flow,tong2024improving} picks a latent $z\sim q(z)$, a conditional path $p_t(x\mid z)$ with $p_0(\cdot\mid z)$, $p_1(\cdot\mid z)$ matching the endpoints in marginal, and a conditional field $u_t(x\mid z)$ generating $p_t(\cdot\mid z)$ through the continuity equation, and minimises
\begin{equation}
\mathcal L_{\mathrm{CFM}}(\theta)=\Ex_{t\sim U[0,1],\,z\sim q,\,x\sim p_t(\cdot\mid z)}\big\|v_\theta(t,x)-u_t(x\mid z)\big\|^2 .
\label{eq:cfm}
\end{equation}
Its gradient equals that of the marginal objective, and the population minimiser is $u_t(x)=\Ex[u_t(x\mid z)\mid x_t=x]$.
The workhorse instance is the Gaussian conditional path $p_t(x\mid z)=\mathcal N(\mu_t(z),\sigma_t(z)^2 I)$, for which the conditional field is
\begin{equation}
u_t(x\mid z)=\dot\mu_t(z)+\frac{\dot\sigma_t(z)}{\sigma_t(z)}\big(x-\mu_t(z)\big),
\label{eq:gausspath}
\end{equation}
\citep[Thm.~3]{lipman2023flow}. \icfm{} is $z=(x_0,x_1)$, $\mu_t=(1-t)x_0+tx_1$, $\sigma_t\equiv\sigma$.

\paragraph{Stream-level flow matching.}
\gpcfm{} \citep{wei2025stream} takes $z$ to be a whole path. A stream $s=\{s_t\}_{t\in[0,1]}$ is a stochastic process with $s_0=x_0$, $s_1=x_1$ and mean-square derivative $\sdot_t$; the per-stream field is $u_t(x\mid s)=\sdot_t$ on the event $s_t=x$, and the loss is
\begin{equation}
\mathcal L_{\mathrm{sCFM}}(\theta)=\Ex_{t,\,s}\big\|v_\theta(t,s_t)-\sdot_t\big\|^2 .
\label{eq:scfm}
\end{equation}
To keep training simulation-free the paper models $s\mid(x_0,x_1)$ as a GP, so that $(s_t,\sdot_t)$ is jointly Gaussian with closed-form mean and covariance obtained by conditioning an auxiliary GP with kernel $c_{11}$ on the anchors $\{(0,x_0),(1,x_1)\}$ (or on $M\ge2$ anchors). Coordinates are treated independently with a shared scalar kernel. All reported experiments use an SE kernel $c_{11}(r)=\alpha^2\exp(-r^2/2\ell^2)$.

\section{What the stream-level objective sees}
\label{sec:reduction}

Write $z$ for the anchors (for the endpoint case $z=(x_0,x_1)$) and $P_t(\cdot\mid z)$ for the joint law of $(s_t,\sdot_t)$ given $z$ at time $t$.

\begin{proposition}[Per-time sufficiency]
\label{prop:reduction}
$\mathcal L_{\mathrm{sCFM}}(\theta)$ and $\nabla_\theta\mathcal L_{\mathrm{sCFM}}(\theta)$ depend on the stream law only through the family $\{P_t(\cdot\mid z)\}_{t\in[0,1]}$: two stream models with the same per-time joint laws have the same objective and the same population gradient at every $\theta$, and the same minimiser $v^\star(t,x)=\Ex[\sdot_t\mid s_t=x]$, which depends on the stream law only through the marginal of $s_t$ and the conditional mean $\Ex[\sdot_t\mid s_t,z]$. Finite-sample gradients agree only when the draws agree: with $n_t=1$ time per stream, common random numbers give identical training trajectories; with $n_t>1$ the cross-time dependence changes the covariance between the $n_t$ summands, so two such models have the same expected gradient but different stochastic gradients, and their SGD trajectories differ.
\end{proposition}
All proofs are in Appendix~\ref{app:proofs}.

Proposition~\ref{prop:reduction} is elementary---the observation that the marginal field is a conditional expectation of \emph{any} process with the right marginals is made in \citet{lipman2024guide}---but it has a sharp consequence for GP streams once combined with mean-square differentiability.

\begin{proposition}[What a Gaussian stream determines]
\label{prop:gaussian}
Let $s\mid z$ be a mean-square differentiable Gaussian process on $\mathbb R^d$ with $\Ex[s_t\mid z]=m_t$, $V_t:=\Var(s_t\mid z)\succ0$, $C_t:=\Cov(s_t,\sdot_t\mid z)$ and $W_t:=\Var(\sdot_t\mid z)$. Then
\begin{align}
\dot V_t&=C_t+C_t^\top, \label{eq:cov}\\
\Ex[\sdot_t\mid s_t,z]&=\mdot_t+\big(\tfrac12\dot V_t-A_t\big)V_t^{-1}(s_t-m_t),
\qquad A_t:=\tfrac12(C_t-C_t^\top), \label{eq:condvel}\\
\Var(\sdot_t\mid s_t,z)&=W_t-C_t^\top V_t^{-1}C_t\;=:\;R_t, \label{eq:resid}
\end{align}
so the path fixes only the symmetric part of $C_t$, and the antisymmetric $A_t$ is free.
Hence (a) the marginal path induced by the stream is the Gaussian conditional path $\mathcal N(m_t,V_t)$, and the term $-A_tV_t^{-1}(s_t-m_t)$ is divergence-free with respect to that path, so the stream fixes the conditional path but fixes the conditional \emph{field} only up to a rotational component; (b) the reachable set of conditional probability paths is therefore exactly the Gaussian conditional paths CFM already parametrises, and for $A_t\equiv0$ the target \eqref{eq:condvel} is exactly the CFM field \eqref{eq:gausspath} of that path; (c) the residual noise in the target does not move the minimiser and only adds $\Ex[\operatorname{tr}R_t]$ to the per-sample target variance; (d) the cross-time covariance $\Cov(s_t,s_{t'}\mid z)$, $t\ne t'$, does not enter $\mathcal L_{\mathrm{sCFM}}$ or its population gradient; it affects only the covariance between the $n_t$ time draws of one stream inside a minibatch.
\end{proposition}

\begin{corollary}[The audited class is two scalar curves]
\label{cor:scalar}
If the stream is built coordinate-wise from one scalar kernel---the construction used throughout \gpcfm, and the only one we audit---then $V_t=v_tI$, $C_t=c_tI$ and $W_t=w_tI$, so $A_t=0$, $c_t=\tfrac12\vdot_t$, and \eqref{eq:condvel}--\eqref{eq:resid} reduce to
\begin{equation}
\Ex[\sdot_t\mid s_t,z]=\mdot_t+\frac{\vdot_t}{2v_t}(s_t-m_t)=\mdot_t+\frac{\dot\sigma_t}{\sigma_t}(s_t-m_t),\qquad
\rho_t^2=w_t-\frac{\vdot_t^2}{4v_t},\qquad \sigma_t:=\sqrt{v_t}.
\label{eq:scalar}
\end{equation}
In this class the whole design space is the pair of scalar curves $(m_t,v_t)$. A general multivariate Gaussian stream keeps one extra degree of freedom, $A_t$, which leaves the path unchanged and adds a rotational component to the target; \gpcfm{} does not use it and we do not claim it is useless.
\end{corollary}

\begin{remark}
Proposition~\ref{prop:gaussian} does not say GP streams are useless; it says they are a \emph{parametrisation} of $(m_t,v_t)$ and nothing more. Any effect of \gpcfm{} relative to \icfm{} must be attributable to (i) the two scalar curves it induces and (ii) the extra target noise $\rho_t$ and the within-minibatch correlation, both of which are variance and not signal. Section~\ref{sec:magnitude} measures all three in the paper's own configuration. (The paper's Algorithm~1 draws one $t$ per stream; the released code draws $n_t=10$ correlated times on the toy benchmarks and $n_t=1$ on images, see Appendix~\ref{app:checks}.)
\end{remark}

We verify \eqref{eq:cov} numerically for the SE bridge used in the paper: the maximal relative discrepancy $\max_t|c_t-\tfrac12\vdot_t|/\max_t|\tfrac12\vdot_t|$ over $\ell\in\{2,1,0.5,0.35,0.25\}$ is $2$--$3\times10^{-8}$ in double precision (Appendix~\ref{app:checks} records two ways this check can falsely fail).

\section{The GP is a constrained chart of \texorpdfstring{$(m_t,v_t)$}{(m,v)}}
\label{sec:chart}

\begin{figure}[t]
\centering
\includegraphics[width=\textwidth]{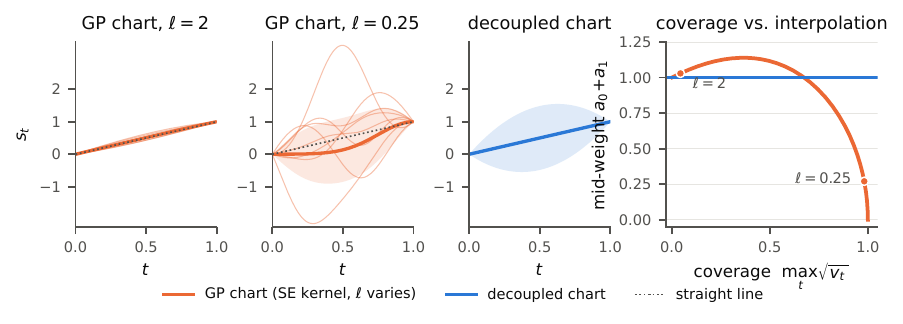}
\caption{\textbf{The SE chart couples coverage and interpolation.} Left two panels: the conditional GP between $x_0=0$ and $x_1=1$ at the paper's $\ell=2$ and at $\ell=0.25$ (mean, $\pm1$\,sd band, eight sample paths). Widening the tube pulls the mean toward the prior: at $\ell=0.25$ the midpoint weight is $0.27$. Third panel: the decoupled chart at the same coverage keeps the straight mean; by Proposition~\ref{prop:reduction} it specifies marginals, not paths, so only the band is drawn. Right: midpoint mean weight against coverage as $\ell$ varies \eqref{eq:sebridge}; the decoupled chart reaches any coverage at weight $1$.}
\label{fig:chart}
\end{figure}

Condition an SE GP with kernel $k(r)=\alpha^2e^{-r^2/2\ell^2}$ on $\{(0,x_0),(1,x_1)\}$. With $\mathbf k_t=(k(t),k(1-t))$ and $K=\big(\begin{smallmatrix}\alpha^2&k(1)\\k(1)&\alpha^2\end{smallmatrix}\big)$,
\begin{equation}
m_t=a_0(t)\,x_0+a_1(t)\,x_1,\quad (a_0,a_1)(t)=\mathbf k_t^\top K^{-1},\qquad
v_t=\alpha^2-\mathbf k_t^\top K^{-1}\mathbf k_t .
\label{eq:sebridge}
\end{equation}

\begin{proposition}[Coverage and interpolation are coupled]
\label{prop:chart}
For fixed $t\in(0,1)$, as $\ell\to0$: $v_t\to\alpha^2$ (maximal coverage) and $a_0(t)+a_1(t)\to0$ (the mean collapses to the prior mean $0$). As $\ell\to\infty$: $v_t\to0$ and $m_t\to$ the linear interpolant. There is no $\ell$ at which $v_t$ is large and $m_t$ interpolates.
\end{proposition}

Figure~\ref{fig:chart} (and Table~\ref{tab:coupling} in Appendix~\ref{app:tables}) quantifies the trade-off. The paper's Table~1 uses $\ell=2$, where the stream is a tube of standard deviation $0.044$ around a mean that overshoots the straight line by $3\%$ at the midpoint. To get a spread comparable to the source scale ($\sigma\approx1$) one needs $\ell\le0.25$, where the midpoint mean weight is $0.27$: the stream from $x_0$ to $x_1$ passes near the origin. The paper's stated recipe (``decrease the SE bandwidth'' to expand coverage) walks straight into this.

\paragraph{A decoupled chart.}
Proposition~\ref{prop:gaussian} licenses writing the two curves directly. We use the straight mean $m_t=(1-t)x_0+tx_1$ and a pinned variance profile $v_t=\sigma_{\max}^2\,(4t(1-t))^2$, whose $t^2$ behaviour at the ends keeps $\vdot_t/(2\sqrt{v_t})$ bounded, with the target \eqref{eq:condvel} and no residual noise. It has one parameter, costs nothing beyond \icfm, and covers the same $(t,x)$ region as the GP at any prescribed $\sigma_{\max}$ without moving the mean.

\section{Reading the released code}
\label{sec:audit}

\begin{figure}[t]
\centering
\includegraphics[width=0.86\textwidth]{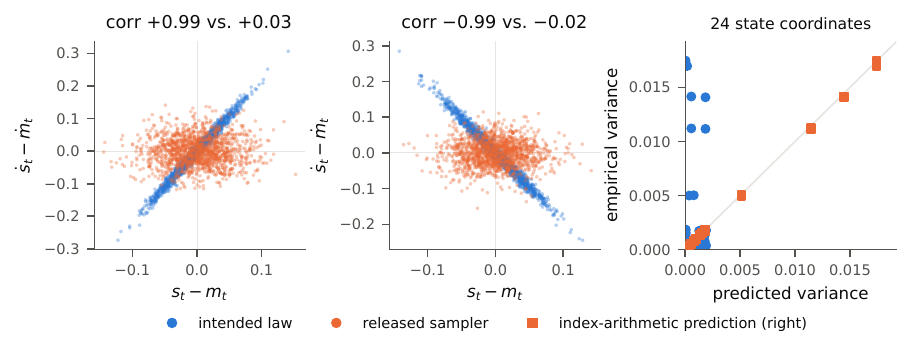}
\caption{\textbf{The released sampler draws state and velocity independently.} Left, centre: $1{,}500$ draws of $(s_t-m_t,\sdot_t-\mdot_t)$ for two streams from the conditional law the paper derives (blue) and from the released \texttt{samp\_x\_dx2} (orange); titles give the intended vs.\ released correlation. Right: empirical variance of every state coordinate against the intended entry $\Sigma_{ii}$ (blue, off the diagonal) and against the value predicted by the index arithmetic of Appendix~\ref{app:bug} (orange squares, on it).}
\label{fig:bug}
\end{figure}

The released repository\footnote{\url{https://github.com/weigcdsb/GP-CFM}, commit \texttt{87fc8ae}.} implements the joint sampler in a function \texttt{samp\_x\_dx2}. For a batch of $n_B$ streams, $n_t$ times and $d$ coordinates it forms the conditional mean \texttt{mu\_new} of shape $(n_B,2n_t,d)$ and the conditional covariance \texttt{Sig\_cond} of shape $(n_B,2n_t,2n_t)$, then
\begin{lstlisting}[basicstyle=\ttfamily\small]
mu_flat = mu_new.view(nB * dim, 2 * nt)
Sig_cond_flat = Sig_cond.repeat(dim, 1, 1)
x_dx = MultivariateNormal(mu_flat, Sig_cond_flat).rsample()
x_dx = x_dx.view(nB, 2 * nt, dim)
\end{lstlisting}
The \texttt{view} reinterprets memory in $(b,i,d)$ order as rows of length $2n_t$; the intended operation is a transpose of the last two axes. Working out the index arithmetic (Appendix~\ref{app:bug}): with $d=2$ and $n_t=10$, coordinate $(b,i,d)$ of the sample is read from row $2b+\lfloor(2i+d)/20\rfloor$, column $(2i+d)\bmod20$. Consequently
(i) the state block $i<n_t$ and the velocity block $i\ge n_t$ come from \emph{different rows}, so $s_t$ and $\sdot_t$ are drawn \emph{independently}; (ii) the variance of coordinate $(i,d)$ is entry $(2i+d,2i+d)$ of the covariance, not $(i,i)$; (iii) row $r$ is paired with \texttt{Sig\_cond[r mod nB]}, so stream $b$ uses the conditional covariance of stream $2b\bmod n_B$.
The mean survives: the same \texttt{view} is applied on the way back, so \texttt{mu\_flat[2b, 2i+d]} lands on $(b,i,d)$. Figure~\ref{fig:bug} (and Table~\ref{tab:bug} in Appendix~\ref{app:tables}) confirms all three effects empirically against the formulas derived in Appendix~\ref{app:bug}. The same \texttt{view}/\texttt{reshape} appears in 5 of 7 notebooks, including \texttt{5\_mnist.ipynb}; the variant in \texttt{3\_intermediate.ipynb} contains a NumPy fallback that indexes \texttt{mu\_new[bb*dim+dd]}---the layout a transpose \emph{would} have produced---which is internal evidence that the reshape is unintended.

What the released code therefore trains on is $s_t\sim\mathcal N(m_t,\tilde v_t)$ with a scrambled $\tilde v_t$ and $\sdot_t\sim\mathcal N(\mdot_t,\tilde w_t)$ \emph{independent} of $s_t$. By Proposition~\ref{prop:reduction} its population target is $\Ex[\mdot_t\mid s_t=x]$, which omits the $\frac{\vdot_t}{2v_t}(x-m_t)$ term of \eqref{eq:condvel} and is therefore not the velocity of any path with the stated marginals. Section~\ref{sec:magnitude} shows the omitted term is $1.7\%$ of the velocity scale in the toy configuration, which is why the code nevertheless ``works'' there; on MNIST the omission is measurable (Section~\ref{sec:mnist}).

We also checked a second fragility: the training loop wraps the sampler in a bare \texttt{try/except: pass}, so a failed positive-definiteness check silently re-uses the previous minibatch. Over $100$ toy runs this happened on $0.06\%$ of steps ($3.1/5000$ on average) and is not a factor.

\section{Experiments}
\label{sec:exp}

All code and per-seed results are in the supplement. Every arm below is trained with the same architecture, optimiser, step budget and number of velocity targets per step as the released code; ``same seed'' means the same initialisation, data order and (where applicable) the same draws of $x_0$ and $t$.

\subsection{Arms}
\label{sec:arms}
\icfm{} is the paper's baseline ($\sigma=0$ on the toy, $10^{-3}$ on CIFAR). \texttt{gp} is the released sampler, verbatim, including the bug and the bare \texttt{except}. \texttt{gpfix} is the law the paper intends: $(s_{t_1..t_{n_t}},\sdot_{t_1..t_{n_t}})$ drawn jointly from the conditional GP with correctly paired mean and covariance (Appendix~\ref{app:impl}). \texttt{mmstoch} matches the GP's per-time law but draws the $n_t$ times independently, isolating cross-time correlation; \texttt{mmdet} keeps $s_t\sim\mathcal N(m_t,v_t)$ with the deterministic target \eqref{eq:condvel}, isolating the residual noise $\rho_t$. On images, \texttt{indep} draws $s_t$ and $\sdot_t$ independently---the structural content of the bug without the index scramble. \texttt{bump} is the decoupled chart of Section~\ref{sec:chart} with $\sigma_{\max}$ matched to the GP's $\max_t\sqrt{v_t}$; \texttt{gpmean\_decvar} and \texttt{straight\_gpvar} cross its mean and variance with the GP's.

\subsection{Magnitudes in the paper's configuration}
\label{sec:magnitude}
For the toy benchmark ($\alpha=1$, $\ell=2$; 2-Gaussian target at $(\pm3,10)$ from $\mathcal N(0,I)$) we evaluate \eqref{eq:sebridge}--\eqref{eq:resid} on a fine grid. Against the transport scale $\Ex\|x_1-x_0\|=10.46$: the widest state spread is $\max_t\sqrt{v_t}=0.044$ ($0.4\%$); the term the bug deletes, $\vdot_t/(2\sqrt{v_t})$, is at most $0.173$ ($1.7\%$); the residual velocity noise is $\max_t\rho_t=0.0128$ ($0.12\%$), so the stream is essentially deterministic given its endpoints; the only quantity of visible size is the mean-path bulge, $a_0(\tfrac12)+a_1(\tfrac12)=1.030$. Figure~1A of the paper, with its visibly wide coverage region, is not the configuration behind its Table~1.

\subsection{The flagship table has no power}
\label{sec:power}

\begin{figure}[t]
\centering
\includegraphics[width=\textwidth]{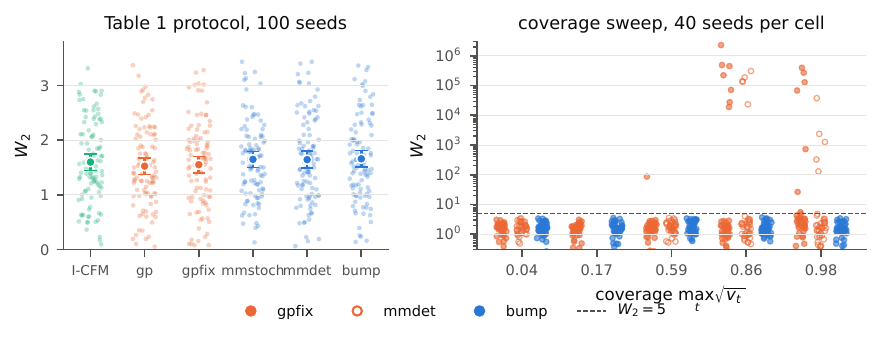}
\caption{\textbf{2-Gaussian benchmark.} Left: $\Wtwo$ of every seed under the paper's Table~1 protocol for the six arms of Section~\ref{sec:arms}, with means and $95\%$ CIs; nothing separates. Right: coverage sweep of Section~\ref{sec:cliff}, log scale. At each level the GP chart (\texttt{gpfix} filled, \texttt{mmdet} hollow: same curves through a different code path) and the decoupled chart (\texttt{bump}) share the same bulk; above $\Wtwo=5$ only orange points appear. Counts are in Table~\ref{tab:cliff}.}
\label{fig:toy}
\end{figure}
We reproduce Table~1 of the paper under its protocol: $N=100$ training targets, 3-hidden-layer MLP of width $64$, $5{,}000$ steps at lr $2\times10^{-3}$ from an arm-specific warm start (itself $5{,}000$ steps at $10^{-3}$), $n_t=10$ targets per pair, $\Wtwo$ between $1{,}000$ generated and $1{,}000$ test samples with dopri5, $100$ seeds. The paper reports \icfm{} $1.54$ and \gpcfm{} $1.51$ with ``SE'' $0.08$, which our replicated per-seed standard deviation of $0.76$ confirms is indeed a standard error.

Figure~\ref{fig:toy} (left) and Table~\ref{tab:toy} (Appendix~\ref{app:tables}) give the result. The reproduction gate passes: \icfm{} $1.60$ vs \texttt{gp} $1.53$, same sign and order as the paper. But the paired difference is $-0.074$ with $95\%$ CI $[-0.257,+0.109]$; the paired standard deviation is $0.93$, so detecting the paper's $0.03$ effect at $80\%$ power and $5\%$ level requires ${\approx}7{,}600$ seeds, and even our own larger point estimate would require ${\approx}1{,}250$. The same holds for every other pair: the six arms---including one with no GP at all---are mutually indistinguishable. This is the expected outcome under Propositions~\ref{prop:gaussian}: the arms differ only in a $0.4\%$ tube, a $0.12\%$ noise and a $3\%$ bulge.

\subsection{Coverage sweep: the cliff and its attribution}
\label{sec:cliff}
Proposition~\ref{prop:chart} predicts that the GP chart fails when coverage is pushed up. We sweep $\ell\in\{2,1,0.5,0.35,0.25\}$ (state spread $0.04$ to $0.98$) and at each level compare \texttt{gpfix} with \texttt{bump} at the same $\sigma_{\max}$, $40$ seeds each, paper protocol. Because the failure is heavy-tailed we report medians, interquartile ranges and the fraction of runs with $\Wtwo>5$ (every functioning model in this benchmark sits near $1.5$; the divergent runs reach $10^4$--$10^5$).

Figure~\ref{fig:toy} (right) and Table~\ref{tab:cliff} show two things. First, the two charts never separate in median quality at any coverage (sign test $p\in[0.08,0.88]$), and the decoupled chart is flat ($1.36$--$1.68$) across a $20\times$ range of coverage: \emph{wider coverage does not help this benchmark}. Second, the tails separate sharply---pooled over the two high-coverage levels the GP chart diverges on $14/80$ runs and the decoupled chart on $0/80$ (Fisher $p=6.6\times10^{-5}$); over all five levels, $15/200$ vs $0/200$ ($p=4.7\times10^{-5}$).

\paragraph{Attribution.}
The divergence could be an artefact of our joint sampler, so \texttt{mmdet} recomputes the same $(m_t,v_t)$ through a different code path (the $2\times2$ per-time conditional of \eqref{eq:sebridge}: no $2n_t\times2n_t$ factorisation, no cross-time draws). It reproduces the cliff---$10/80$, statistically identical to \texttt{gpfix} ($p=0.51$) and distinct from \texttt{bump} ($p=1.5\times10^{-3}$)---and guarding the $\vdot_t/2\sqrt{v_t}$ evaluation near the pinned endpoints (Appendix~\ref{app:checks}) leaves the counts unchanged. It could also be a property of the variance profile rather than the mean that Proposition~\ref{prop:chart} blames, so we cross the two curves (Table~\ref{tab:iso}, Appendix~\ref{app:tables}): keeping the GP mean and swapping in the decoupled variance leaves the cliff intact ($11/80$ vs $14/80$, $p=0.66$), while keeping the GP variance and straightening the mean removes it completely ($0/80$; $p=6.6\times10^{-5}$ against the full GP, $p=1.00$ against the decoupled chart). Divergence tracks the mean and is insensitive to the variance---what Proposition~\ref{prop:chart} predicts, since what breaks at small $\ell$ is that $m_t$ stops interpolating.

\paragraph{Cost.} Wall-clock per $5{,}000$-step run on one CPU thread: \texttt{gp} $439$\,s, \texttt{gpfix} $173$\,s, \texttt{mmstoch} $81$\,s, \texttt{mmdet} $80$\,s, \texttt{bump} $60$\,s, \icfm{} $51$\,s---the joint sampler costs $2$--$5\times$ writing the two curves down, for the same median result.

\subsection{MNIST}
\label{sec:mnist}

We follow \texttt{5\_mnist.ipynb} line by line: UNet with $32$ channels and one residual block \citep{nichol2021improved}, Adam at $2\times10^{-4}$, batch $128$, $5$ epochs, $\sigma=0$, GP with $\alpha=1$, $\ell=5$, $n_t=1$; FID and KID computed with the repository's own \texttt{metric/Fid\_score.py} between $1{,}280$ generated and $1{,}280$ test images. Within a seed, all arms share the initialisation, the data order and the draws of $x_0$ and $t$, so the comparison is paired at the level of the training trajectory.

\begin{table}[t]
\centering
\caption{MNIST, paper protocol, $20$ seeds per arm; paired differences over the same seeds (Wilcoxon signed-rank). The released method is worse than \icfm; every correct implementation of the intended law is equivalent to \icfm{} within the paper's claimed $0.95$ FID effect (TOST $p\le0.006$); removing the bug in any of three ways recovers $2.2$--$2.3$ FID. The paper reports \icfm{} $44.50\,(0.18)$ and \gpcfm{} $43.55\,(0.13)$.}
\label{tab:mnist}
\small
\resizebox{\textwidth}{!}{%
\begin{tabular}{l rr rr rr}
\toprule
arm & FID & KID & $\Delta$FID vs \icfm{} [$95\%$ CI] & $p$ & $\Delta$FID vs \texttt{gp} [$95\%$ CI] & $p$ \\
\midrule
\icfm & $33.33\pm0.70$ & $0.0214$ & --- & & $-2.56$ & \\
\texttt{gp} (released) & $35.90\pm0.78$ & $0.0338$ & $+2.56$ $[+1.71,+3.42]$, wins $1/20$ & $<10^{-3}$ & --- & \\
\texttt{gpfix} & $33.70\pm0.74$ & $0.0217$ & $+0.37$ $[-0.04,+0.78]$ & $0.11$ & $-2.19$ $[-3.04,-1.34]$ & $<10^{-3}$ \\
\texttt{indep} & $33.61\pm0.70$ & $0.0216$ & $+0.28$ $[-0.09,+0.64]$ & $0.20$ & $-2.29$ $[-3.11,-1.46]$ & $<10^{-3}$ \\
\texttt{mmdet} & $33.67\pm0.73$ & $0.0218$ & $+0.34$ $[-0.03,+0.70]$ & $0.10$ & $-2.23$ $[-3.08,-1.37]$ & $<10^{-3}$ \\
\bottomrule
\end{tabular}}
\end{table}

Table~\ref{tab:mnist} (drawn per seed in Figure~\ref{fig:mnist}) reverses the paper's MNIST conclusion. The released \gpcfm{} is worse than \icfm{} by $2.56$ FID (CI $[1.71,3.42]$, $1$ win in $20$ paired seeds), and its KID is $1.6\times$ that of every other arm. The three correct implementations---joint law, independent state/velocity, deterministic target---sit within $0.4$ FID of \icfm{}, which we test as equivalence rather than as a failure to reject: with the margin set to the effect the paper claims ($44.50-43.55=0.95$ FID), two one-sided tests reject a difference of that size in all three ($p=0.006$, $0.0008$, $0.002$). They are not identical to \icfm{}---point estimates $+0.28$ to $+0.37$, two of three $90\%$ intervals excluding zero---but if anything slightly \emph{worse}, and nowhere near a $0.95$ gain, as Proposition~\ref{prop:gaussian} predicts for a $0.4\%$-scale perturbation. And the bug is not neutral on images: relative to the released code, \texttt{indep} (which keeps state and velocity independent but removes the pixel misindexing and wrong-stream covariance) recovers $2.3$ FID, so the damage is in the scrambled variances, not in the independence per se. Our absolute FIDs ($33$) are below the paper's ($44$) although training and the FID routine are the paper's (it draws its $1{,}280$ reference images with \texttt{shuffle=True}, we fix them); the offset affects all arms equally.

\subsection{Does the cliff appear on images?}
\label{sec:mnsweep}
The toy cliff concerns a regime the paper never enters: its MNIST kernel ($\alpha=1$, $\ell=5$) gives $\max_t\sqrt{v_t}=0.0071$---$0.7\%$ of the pixel range---at a midpoint mean weight of $1.005$. We repeat the sweep on MNIST, pushing $\ell$ down as the paper recommends, with the same two charts and ten seeds per cell.

Table~\ref{tab:mnsweep} (Appendix~\ref{app:tables}) does not reproduce the toy result, and we report it as it came out. Neither chart diverges on images---one \texttt{bump} run in $100$ exceeds FID $100$, no GP run does---and at matched coverage the GP chart is the \emph{better} of the two, by $3$ to $15$ FID, the opposite ordering to Table~\ref{tab:iso}. The coupling of Proposition~\ref{prop:chart} is analytic and holds everywhere, but whether it is harmful is benchmark-dependent, ranging from divergence on the toy to an advantage here.

Two things do transfer. At the paper's own setting the two charts are indistinguishable ($+0.41$ FID): with $0.7\%$ coverage there is nothing for either to do. And the recipe does not pay---no level is significantly better than the paper's ($\ell=1$ is the best candidate, $-2.74$ FID, $95\%$ CI $[-6.22,+0.75]$), while past $\max_t\sqrt{v_t}\approx0.6$ both degrade significantly: \texttt{gpfix} by $+16.8$ FID at $\ell=0.25$ and \texttt{bump} by $+9.9$ already at $\ell=0.5$ (both $p=0.002$). On the paper's own image benchmark, ``decrease the SE bandwidth to expand the coverage region'' has no upside and a large downside.

\subsection{CIFAR-10}
\label{sec:cifar}
The paper's Table~4 reports FID $3.75$ (\icfm), $3.74$ (OT-CFM), $3.62$ (\gpcfm) and $3.75$ (GP-OT-CFM) with standard errors of $0.006$--$0.009$. Those standard errors come from re-evaluating FID $20$ times on \emph{one} trained model per method; no training-seed variance is measured, so the $0.13$ gap between \gpcfm{} and \icfm{} is one run against one run. We train all three arms for five seeds each in the configuration the paper cites \citep{tong2024improving} ($128$-channel UNet, $400$k steps, EMA; details in Appendix~\ref{app:impl}), with FID on $50$k samples from the reference \texttt{clean-fid} pipeline. No CIFAR code or kernel parameters are released; we use the MNIST notebook's ($\alpha=1$, $\ell=5$).

Table~\ref{tab:cifar} (Appendix~\ref{app:tables}) makes three points. First, our \icfm{} ($3.63\pm0.09$) matches the paper's ($3.75$), and the intended GP law ($3.68\pm0.07$) is equivalent to it within the paper's own margin: the paired difference is $+0.050$ FID and TOST at the $0.13$ margin gives $p=0.017$ paired, $p=0.082$ unpaired. Second, the paper's \gpcfm{} number ($3.62$) lies inside the \icfm{} seed range $[3.55,3.79]$: with a single-run sd of $0.094$, the difference of two single runs has sd $0.133$, so the reported $0.13$ gap is one such unit. Third, the released sampler does not reproduce $3.62$: it is worse than \icfm{} by $1.67$ FID on every seed ($+1.49$ to $+1.76$), from $100$k steps on, with a higher training loss throughout ($0.177$ vs $0.169$; the scrambled variances are irreducible target noise). No CIFAR code was released, so which sampler produced the paper's $3.62$ is unknown---we report that the released one does not reach it and the intended law does not beat \icfm{}, without attributing the paper's number to the bug.

\section{Related work}
\label{sec:related}
\paragraph{Flow matching and its design space.}
\citet{lipman2023flow} introduced flow matching with Gaussian conditional paths and \eqref{eq:gausspath}; \citet{tong2024improving} generalised the conditioning to arbitrary $z$ and couplings; \citet{albergo2023building,albergo2023stochastic} introduced stochastic interpolants, whose benefit is likewise explained by the induced marginal. \citet{lipman2024guide} note that the marginal field is a conditional expectation of any process with the right marginals; Proposition~\ref{prop:reduction} is the stream-level instance, and our contribution is its consequence for GP streams (Proposition~\ref{prop:gaussian}) and the coupling failure of the SE chart (Proposition~\ref{prop:chart}). \citet{shaul2023kinetic} characterise kinetic-optimal Gaussian paths; our decoupled chart is a point in that family. Path and coupling design remains the field's main lever---straightened couplings and their training \citep{liu2023rectified,pooladian2023multisample,lee2024improving,esser2024scaling}, bridge and Schr\"odinger-bridge matching \citep{shi2023dsbm}, data-dependent interpolants \citep{albergo2024couplings}, metric-aware and manifold paths \citep{kapusniak2024metric,chen2024riemannian}, least-action and scheduled paths \citep{du2026lagrangian,bondar2026velocity}, the design-space views of \citet{karras2022elucidating} and \citet{ma2024sit}, few-step objectives \citep{song2023consistency,frans2025shortcut,geng2025meanflow}---all of it acting on the coupling, the geometry, or $(m_t,v_t)$, inside the space Proposition~\ref{prop:gaussian} identifies. \citet{ma2024sit} is closest in spirit: it ablates interpolant choices at scale and finds the path worth tuning---consistent with our reading, in which the lever is $(m_t,v_t)$ and the only question is how one parametrises it.
\citet{closedform2025} find stochastic and closed-form CFM targets nearly equivalent and \citet{stablevelocity2026} analyse the variance of single-sample conditional velocities; both agree with Proposition~\ref{prop:gaussian}(c), and the deterministic target is not our contribution.

\section{Limitations}
\label{sec:limits}
The positive result is about controllability and cost, not accuracy: at the paper's own operating point no arm beats any other in median quality on any benchmark, and we report every arm we ran, including those where the audited method wins on points. The cliff is a toy result---on MNIST the same sweep gives no divergence and reverses the ordering of the two charts (Section~\ref{sec:mnsweep})---so Proposition~\ref{prop:chart} bounds what the chart \emph{can} do, not what it always does; we did not sweep coverage on CIFAR-10, which also uses assumed kernel parameters. The reduction is stated for the latent $z$ the algorithm conditions on, so it covers the $M>2$ anchors of the paper's Section~4.2 (untested) but not random anchors \emph{marginalised out} of $z$; the ``two curves'' reading is Corollary~\ref{cor:scalar}, specific to the coordinate-wise shared-kernel construction. The MNIST FID offset to the paper is unexplained.

\section{Conclusion}
For the Gaussian streams \gpcfm{} builds, the stream-level view adds nothing to a conditional path: the objective sees two scalar curves, mean-square differentiability fixes their coupling, and the rest is variance. The GP ties the curves together, so the method's own recipe for coverage collapses its mean---divergence on the toy, no benefit on images. And \gpcfm's record does not survive its own protocol: the sampler is wrong, the flagship table lacks the power to see its claim, the MNIST gain reverses sign under pairing, and on CIFAR-10 the released sampler never reaches the reported number.

\clearpage
\section*{Reproducibility statement}
All experiments are reproducible from the supplementary code. The 2-Gaussian experiments (\texttt{derisk.py}) reproduce the released notebook's data, network, optimiser, warm start and $\Wtwo$ evaluation and add the arms of Section~\ref{sec:arms} behind a single flag; the sampler test (\texttt{test2.py}) and the draws behind Figure~\ref{fig:bug} (\texttt{dump\_bug.py}) call the released \texttt{samp\_x\_dx2} verbatim. MNIST (\texttt{mnist\_derisk.py}) and CIFAR-10 (\texttt{cifar\_derisk.py}) fix, per seed, the generators for initialisation, data order, and the shared $x_0$ and $t$ draws, so every arm can be re-paired exactly; the FID routines are the released repository's \texttt{metric/Fid\_score.py} and \texttt{clean-fid} in the reference configuration. All hyper-parameters are stated in Section~\ref{sec:exp} and Appendix~\ref{app:impl}; the per-seed result files behind every table and the figure script (\texttt{figs/make\_figs.py}, which reads them and hard-codes no number) are included. Proofs are in Appendix~\ref{app:proofs}; the index arithmetic behind the code finding is in Appendix~\ref{app:bug}; the numerical checks that guard the theory are in Appendix~\ref{app:checks}.

\bibliography{gpcfm_audit_preprint}
\bibliographystyle{preprint}

\appendix
\section{Proofs}
\label{app:proofs}
\paragraph{Proposition~\ref{prop:reduction}.}
The integrand of \eqref{eq:scfm} is a function of $(t,s_t,\sdot_t)$ only; the expectation over $s$ therefore factors through $P_t(\cdot\mid z)$ for each $t$. The rest is the standard least-squares argument.

\paragraph{Proposition~\ref{prop:gaussian}.}
\eqref{eq:cov}: mean-square differentiability gives $\frac{d}{dt}\Ex[(s_t-m_t)(s_t-m_t)^\top]=\Ex[(\sdot_t-\mdot_t)(s_t-m_t)^\top]+\Ex[(s_t-m_t)(\sdot_t-\mdot_t)^\top]=C_t^\top+C_t$. Only the symmetric part of $C_t$ is therefore determined by $V_t$; write $C_t=\tfrac12\dot V_t+A_t$ with $A_t$ antisymmetric.
\eqref{eq:condvel}--\eqref{eq:resid}: Gaussian conditioning, $\Ex[\sdot_t\mid s_t]=\mdot_t+\Cov(\sdot_t,s_t)V_t^{-1}(s_t-m_t)$ with $\Cov(\sdot_t,s_t)=C_t^\top=\tfrac12\dot V_t-A_t$, and $\Var(\sdot_t\mid s_t)=W_t-C_t^\top V_t^{-1}C_t$.
(a) The marginal of $s_t$ given $z$ is $\mathcal N(m_t,V_t)$ by construction. For the divergence-free claim, let $p=\mathcal N(m_t,V_t)$ and $u(x)=B(x-m_t)$ with $B=-A_tV_t^{-1}$; then $\nabla\!\cdot\!(pu)=p\,[\operatorname{tr}B-(x-m_t)^\top V_t^{-1}B(x-m_t)]$, and $V_t^{-1}B=-V_t^{-1}A_tV_t^{-1}$ is antisymmetric (as $V_t$ is symmetric), so both terms vanish identically. Hence the two fields $\mdot_t+\tfrac12\dot V_tV_t^{-1}(\cdot-m_t)$ and \eqref{eq:condvel} generate the same path.
(b) Follows from (a) and Proposition~\ref{prop:reduction}; with $A_t=0$, \eqref{eq:condvel} is \eqref{eq:gausspath} with $(\mu_t,\Sigma_t)=(m_t,V_t)$.
(c) Adding zero-mean noise independent of $(s_t,z)$ to a least-squares target leaves the minimiser unchanged and adds its trace to the target variance.
(d) The minibatch estimator draws $z$ and then $n_t$ times $t_1,\dots,t_{n_t}$ from the same stream; the expectation of each summand depends on $P_{t_j}$ only, while the covariance between summands depends on the joint law across $t_j,t_k$.

\paragraph{Corollary~\ref{cor:scalar}.} With $V_t=v_tI$ etc.\ every matrix in Proposition~\ref{prop:gaussian} is a multiple of $I$, hence symmetric, so $A_t=0$ and \eqref{eq:cov} reads $2c_t=\vdot_t$; substituting into \eqref{eq:condvel}--\eqref{eq:resid} gives \eqref{eq:scalar}.

\paragraph{Proposition~\ref{prop:chart}.}
$k(t),k(1-t)\to0$ as $\ell\to0$ for $t\in(0,1)$, so $\mathbf k_t\to0$; $K^{-1}$ stays bounded. The $\ell\to\infty$ limit is the standard linear limit of the SE interpolant.

\section{Index arithmetic of the released sampler}
\label{app:bug}
Let \texttt{mu\_new} have shape $(n_B,2n_t,d)$, stored contiguously, so that entry $(b,i,k)$ sits at flat offset $f=b\cdot2n_td+i\cdot d+k$. \texttt{view(nB*dim, 2*nt)} maps offset $f$ to row $r=\lfloor f/2n_t\rfloor$, column $j=f\bmod2n_t$. For $d=2$, $n_t=10$: $f=40b+2i+k$, so $r=2b+\lfloor(2i+k)/20\rfloor$ and $j=(2i+k)\bmod20$. For $i<10$ (state block) $r=2b$; for $i\ge10$ (velocity block) $r=2b+1$. Hence:
\begin{itemize}
\item State and velocity of the same $(b,k)$ are in rows $2b$ and $2b+1$, sampled as independent multivariate normals.
\item Within row $2b$, the coordinate $(i,k)$ occupies column $2i+k$; its variance is \texttt{Sig[2i+k, 2i+k]}, which for $2i+k\ge10$ is a \emph{velocity} variance.
\item \texttt{Sig\_cond.repeat(dim,1,1)} makes row $r$ use \texttt{Sig\_cond[r mod nB]}, i.e.\ stream $2b\bmod n_B$.
\item The reverse \texttt{view(nB, 2*nt, dim)} applies the identical map, so the mean is placed correctly: \texttt{mu\_flat[2b, 2i+k]} $=$ \texttt{mu\_new[b,i,k]}.
\end{itemize}
The ``bug formula'' column of Table~\ref{tab:bug} (Appendix~\ref{app:tables}) is $\texttt{Sig\_cond}_{(2b\bmod n_B)}[2i+k,2i+k]$; all $24$ coordinates of the $n_B=4$, $n_t=3$, $d=2$ test agree with it to three decimals, and the four intended correlations $\pm0.83$--$0.99$ are measured at $0.00\pm0.02$ ($6{,}000$ draws, s.e.\ $0.013$).

The nt$=1$ variant used on images (\texttt{5\_mnist.ipynb}) has $f=2b d+i d+k$ with $i\in\{0,1\}$; the same map pairs the state of pixel $k$ with the velocity of a different pixel and draws them from different rows, so state--velocity independence holds there too.

\section{Numerical checks and two ways to fool them}
\label{app:checks}
\paragraph{Algorithm vs.\ code.} The paper's Algorithm~1 draws one $t$ per stream, in which case cross-time covariance is never touched at all. The released code draws $n_t=10$ correlated times per stream on the toy benchmarks and $n_t=1$ on images. The statement above is written for the general case: cross-time structure is variance-only, not unused.

We assert $c_t=\tfrac12\vdot_t$ \eqref{eq:cov} on a grid before every run. Two versions of this assertion failed spuriously and are recorded because the paper audits others' numerics.
(i) In single precision with $v_t\sim10^{-3}$ the relative discrepancy was $0.16$; in double precision it is $10^{-11}$ pointwise. All per-time statistics are computed in double precision.
(ii) A \emph{pointwise} relative error $|c_t-\tfrac12\vdot_t|/|\tfrac12\vdot_t|$ reached $1.0$ at $t=\tfrac12$, where by symmetry $\vdot_t\equiv0$ and both sides are $10^{-14}$; the ratio is round-off. The assertion uses $\max_t|c_t-\tfrac12\vdot_t|/\max_t|\tfrac12\vdot_t|$, which is $2$--$3\times10^{-8}$ for every $\ell$ used.
Near the pinned endpoints $v_t\sim t^2\to0$ and the slope $\vdot_t/(2\sqrt{v_t})$ in \eqref{eq:condvel} is $0/0$ in floating point (observed $10^{6}$); analytically it tends to $\mathrm{sign}(\vdot_t)\sqrt{w_t}$ and $\rho_t\to0$. We use these limits wherever $v_t<10^{-9}$ and assert the slope stays below $10\max_t\sqrt{w_t}$. Re-running the \texttt{mmdet} attribution arm with this guard left every count in Table~\ref{tab:cliff} unchanged.

\section{Additional tables and figures}
\label{app:tables}
\begin{table}[h]
\centering
\caption{MNIST coverage sweep, ten seeds per cell, paper protocol otherwise. The cliff does \emph{not} transfer: neither chart diverges and the GP chart is the better of the two at matched coverage, the opposite ordering to the toy. What does transfer: no level beats the paper's own, and past $\max_t\sqrt{v_t}\approx0.6$ both charts degrade.}
\label{tab:mnsweep}
\small
\resizebox{\textwidth}{!}{%
\begin{tabular}{rr cc c}
\toprule
$\max_t\sqrt{v_t}$ & $\ell$ & \texttt{gpfix} (GP chart) & \texttt{bump} (decoupled) & paired \texttt{gpfix}$-$\texttt{bump} [$95\%$ CI] \\
\midrule
0.0071 & 5 (paper) & $33.53\pm1.10$ & $33.12\pm1.05$ & $+0.41$ $[-0.07,+0.89]$ \\
0.1745 & 1    & $30.80\pm1.01$ & $34.26\pm1.34$ & $-3.46$ $[-6.66,-0.27]$ \\
0.5933 & 0.5  & $33.91\pm1.39$ & $43.04\pm3.14$ & $-9.13$ $[-15.79,-2.47]$ \\
0.8628 & 0.35 & $42.02\pm4.04$ & $57.06\pm5.88$ & $-15.04$ $[-29.87,-0.21]$ \\
0.9815 & 0.25 & $50.36\pm4.49$ & $59.96\pm6.96$ & $-9.60$ $[-25.58,+6.39]$ \\
\bottomrule
\end{tabular}}
\end{table}

\begin{table}[h]
\centering
\caption{Which curve causes the cliff? Divergent runs ($\Wtwo>5$) out of $40$ seeds per cell at the two high-coverage levels, with the mean and variance curves crossed. Replacing the GP's variance changes nothing; replacing its mean removes the cliff entirely. \texttt{gpfix} and \texttt{mmdet} are independent implementations of the same (GP, GP) cell. Medians and the full five-level sweep, where no arm diverges below coverage $0.6$, are in Table~\ref{tab:cliff}.}
\label{tab:iso}
\small
\begin{tabular}{llccc}
\toprule
mean $m_t$ & variance $v_t$ & arm & $\ell=0.35$ & $\ell=0.25$ \\
\midrule
GP & GP & \texttt{gpfix} & 7/40 & 7/40 \\
GP & GP & \texttt{mmdet} & 5/40 & 5/40 \\
GP & decoupled & \texttt{gpmean\_decvar} & 5/40 & 6/40 \\
straight & GP & \texttt{straight\_gpvar} & \textbf{0/40} & \textbf{0/40} \\
straight & decoupled & \texttt{bump} & \textbf{0/40} & \textbf{0/40} \\
\bottomrule
\end{tabular}
\end{table}

\begin{table}[h]
\centering
\caption{CIFAR-10, $400$k steps, FID with $50$k samples (lower is better); five training seeds per arm; the paper's single-run numbers for reference.}
\label{tab:cifar}
\small
\begin{tabular}{l ccccc c c}
\toprule
arm & seed 0 & seed 1 & seed 2 & seed 3 & seed 4 & mean $\pm$ sd & paper (one run) \\
\midrule
\icfm & 3.79 & 3.64 & 3.61 & 3.55 & 3.58 & $3.63\pm0.09$ & 3.75 \\
\texttt{gp} (released) & 5.29 & 5.37 & 5.21 & 5.30 & 5.34 & $5.30\pm0.06$ & 3.62 \\
\texttt{gpfix} (intended law) & 3.79 & 3.61 & 3.68 & 3.65 & 3.67 & $3.68\pm0.07$ & --- \\
\bottomrule
\end{tabular}
\end{table}

\begin{table}[h]
\centering
\caption{Coverage sweep on the 2-Gaussian benchmark ($40$ seeds per cell). Medians never separate (paired sign test $p\ge0.08$ at every level); the tails do. Right block: attribution arm \texttt{mmdet}, which uses the GP's own $(m_t,v_t)$ through a $2\times2$ per-time computation and never touches the joint sampler.}
\label{tab:cliff}
\small
\resizebox{\textwidth}{!}{%
\begin{tabular}{rr rr rr rr}
\toprule
\multirow{2}{*}{$\max_t\sqrt{v_t}$} & \multirow{2}{*}{$\ell$} & \multicolumn{2}{c}{\texttt{gpfix} (GP chart)} & \multicolumn{2}{c}{\texttt{bump} (decoupled)} & \multicolumn{2}{c}{\texttt{mmdet} (GP curves, no sampler)} \\
 & & median [IQR] & diverged & median [IQR] & diverged & median & diverged \\
\midrule
0.044 & 2.00 & 1.63 [0.93, 1.88] & 0/40 & 1.59 [1.12, 2.08] & 0/40 & 1.65 & 0/40 \\
0.175 & 1.00 & 1.54 [0.93, 1.81] & 0/40 & 1.57 [1.14, 2.00] & 0/40 & --- & --- \\
0.593 & 0.50 & 1.52 [1.18, 2.12] & 1/40 & 1.68 [1.21, 2.56] & 0/40 & 1.77 & 0/40 \\
0.863 & 0.35 & 1.60 [0.94, 2.52] & \textbf{7/40} & 1.46 [1.06, 2.01] & 0/40 & 1.69 & \textbf{5/40} \\
0.982 & 0.25 & 2.11 [0.85, 3.39] & \textbf{7/40} & 1.36 [1.14, 1.89] & 0/40 & 1.62 & \textbf{5/40} \\
\bottomrule
\end{tabular}}
\end{table}

\begin{table}[h]
\centering
\caption{2-Gaussian benchmark, $100$ seeds per arm, paper protocol. Paired differences use the same seed for initialisation and evaluation noise. No pair of arms is distinguishable; the paper's reported gap ($-0.03$) lies well inside every interval.}
\label{tab:toy}
\small
\begin{tabular}{l rrr r}
\toprule
arm & mean $\Wtwo$ & sd & se & paired diff vs.\ \texttt{gp} [$95\%$ CI] \\
\midrule
\icfm & 1.601 & 0.765 & 0.077 & $+0.074$ $[-0.109,+0.257]$ \\
\texttt{gp} (released) & 1.526 & 0.757 & 0.076 & --- \\
\texttt{gpfix} & 1.553 & 0.780 & 0.078 & $+0.027$ $[-0.148,+0.201]$ \\
\texttt{mmstoch} & 1.651 & 0.741 & 0.074 & $+0.124$ $[-0.051,+0.300]$ \\
\texttt{mmdet} & 1.645 & 0.791 & 0.079 & $+0.119$ $[-0.073,+0.311]$ \\
\texttt{bump} (no GP) & 1.661 & 0.772 & 0.077 & $+0.135$ $[-0.054,+0.323]$ \\
\bottomrule
\end{tabular}
\end{table}

\begin{figure}[h]
\centering
\includegraphics[width=0.48\textwidth]{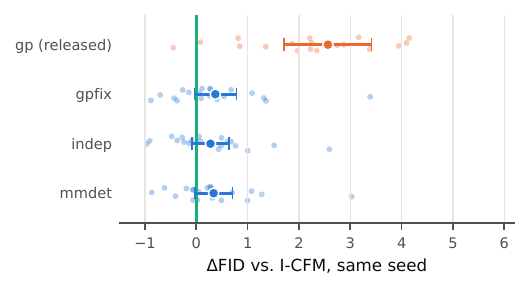}
\caption{\textbf{MNIST, paired by seed.} Per-seed FID difference to \icfm{} (light points; green line is \icfm) with mean and $95\%$ CI. The released code is worse on $19/20$ seeds; the three correct implementations of the intended law sit on \icfm{}.}
\label{fig:mnist}
\end{figure}

\begin{table}[h]
\centering
\caption{The SE chart couples coverage and interpolation \eqref{eq:sebridge}. $\max_t\sqrt{v_t}$ is the widest state spread; $a_0+a_1$ at $t=\tfrac12$ equals $1$ for a straight interpolant; the last column is the maximal deviation of $m_t$ from the straight line per unit endpoint separation.}
\label{tab:coupling}
\small
\begin{tabular}{r rrr}
\toprule
$\ell$ & $\max_t\sqrt{v_t}$ & $a_0(\tfrac12)+a_1(\tfrac12)$ & $\max_t|m_t-\mathrm{lin}_t|$ \\
\midrule
4.00 & 0.011 & 1.008 & 0.005 \\
2.00 & 0.044 & 1.030 & 0.018 \\
1.00 & 0.175 & 1.099 & 0.062 \\
0.50 & 0.593 & 1.069 & 0.104 \\
0.25 & 0.982 & 0.271 & 0.365 \\
0.10 & 1.000 & 0.000 & 0.706 \\
\bottomrule
\end{tabular}
\end{table}

\begin{table}[h]
\centering
\caption{The released sampler does not draw from the law it derives ($n_B=4$, $n_t=3$, $d=2$, $6{,}000$ draws per stream). Left: state--velocity correlation at $t_1$. Right: variance of selected coordinates vs.\ the value predicted by the index arithmetic of Appendix~\ref{app:bug} and the intended value; all $24$ coordinates agree with the bug formula to three decimals.}
\label{tab:bug}
\small
\begin{tabular}{c rr c ccc rrr}
\toprule
\multirow{2}{*}{$b$} & \multicolumn{2}{c}{$\mathrm{corr}(s_{t_1},\sdot_{t_1})$} & & \multirow{2}{*}{$b$} & \multirow{2}{*}{$i$} & \multirow{2}{*}{$d$} & \multicolumn{3}{c}{$\Var(s^{(d)}_{t_i})$} \\
 & empirical & intended & & & & & empirical & bug formula & intended \\
\midrule
0 & $+0.003$ & $-0.989$ & & 0 & 1 & 1 & 0.0050 & 0.0051 & 0.0008 \\
1 & $+0.024$ & $+0.991$ & & 0 & 2 & 0 & 0.0112 & 0.0114 & 0.0006 \\
2 & $+0.005$ & $+0.950$ & & 1 & 2 & 0 & 0.0172 & 0.0174 & 0.0002 \\
3 & $-0.014$ & $-0.834$ & & 3 & 2 & 0 & 0.0176 & 0.0174 & 0.0000 \\
\bottomrule
\end{tabular}
\end{table}

\section{Implementation details}
\label{app:impl}
\paragraph{\texttt{gpfix}.} For each stream we form the $2n_t\times2n_t$ conditional covariance exactly as the released code does, symmetrise, take an eigendecomposition, clamp negative eigenvalues to zero, and draw $L\varepsilon$ with $L=U\Lambda^{1/2}$; the mean of shape $(n_B,2n_t,d)$ is transposed to $(n_Bd,2n_t)$ and each row paired with its own stream's factor. No positive-definiteness exception can occur.
\paragraph{Toy protocol.} Data, network, optimiser and evaluation are the released notebook's (\texttt{1\_varRed.ipynb}); the warm start is arm-specific as in the notebook's 100-seed loop; $\Wtwo$ uses POT's exact EMD with squared Euclidean cost between $1{,}000$ generated samples (dopri5, atol $=$ rtol $=10^{-4}$) and $1{,}000$ held-out targets drawn with the notebook's seed.
\paragraph{Equivalence tests.} Where we claim an arm matches \icfm{} we run two one-sided tests (TOST) with the margin declared in advance as the effect the audited paper claims: $44.50-43.55=0.95$ FID on MNIST and $3.75-3.62=0.13$ FID on CIFAR-10. The reported $p$ is $\max$ of the two one-sided $p$-values, over the paired differences on the $20$ (MNIST) and $5$ (CIFAR-10) shared seeds; the unpaired Welch version is also reported for CIFAR-10.

\paragraph{Run-to-run nondeterminism.} Re-running the same arm and seed on a different GPU changes MNIST FID by $+0.02$ on average and at most $0.33$ over ten seeds (cuDNN convolution backward is nondeterministic by default). This bounds the noise in our paired image comparisons: it is an order of magnitude below the effects we report for the released sampler and comparable to the $\pm0.4$ equivalence margins, which is why those are reported as equivalence tests rather than as null results.

\paragraph{Power calculation.} With paired standard deviation $s_d$ and effect $\delta$, the two-sided $5\%$, $80\%$-power sample size is $n\approx(2.8\,s_d/\delta)^2$; $s_d=0.934$ gives $7{,}595$ for $\delta=0.03$ and $1{,}248$ for $\delta=0.074$.
\paragraph{MNIST.} Per seed, one generator drives the initialisation, one the data order, one the draws of $x_0$ and $t$ (shared by all arms) and one the arm-specific noise; FID and KID call the repository's \texttt{metric/Fid\_score.py} and \texttt{torchmetrics} KID (subset size $40$) on $1{,}280$ generated vs.\ $1{,}280$ fixed test images.
\paragraph{CIFAR-10.} Training follows \texttt{train\_cifar10.py} of the reference implementation with the paper's $\sigma=10^{-3}$ and $10^{-6}$ jitter; within a seed all arms share initialisation, data order, $x_0$ and $t$. FID: \texttt{clean-fid} in \texttt{legacy\_tensorflow} mode against CIFAR-10 train statistics, $50$k samples from the EMA model integrated with dopri5 at tolerance $10^{-5}$ (intermediate checkpoints: $10$k samples). Full configuration: UNet with $128$ channels and $2$ residual blocks, Adam at $2\times10^{-4}$ with $5$k warm-up, gradient clipping at $1$, EMA $0.9999$, batch $128$, random horizontal flips, $400$k steps. Wall-clock $\approx26$--$40$\,h per run on one A6000 depending on load. Intermediate FIDs ($10$k samples) at $100$k/$200$k/$300$k steps: \icfm{} $8.22/6.59/6.08$, $7.67/6.27/5.77$, $7.75/6.21/5.81$ (seeds 0--2); released \texttt{gp} $9.05/7.95/7.49$, $9.13/7.86/7.47$, $8.89/7.77/7.36$; \texttt{gpfix} $8.12/6.49/6.01$.

\section*{AI use statement}
Parts of this paper's text, code and analysis were produced with the assistance of generative AI. The authors take full responsibility for the final content of this paper, including any text produced with the assistance of generative AI.

\end{document}